# Research and Implementation of Global Path Planning for Unmanned Surface Vehicle Based on Electronic Chart

[1]Yanlong Wang, [1]Xu Liang, [1]Baoan Li, [2]Xuemin Yu

[1]School of Automation Science and Electrical Engineering, Beihang University, Beijing, China

[2]Institute of Information Engineering, Chinese Academy of Sciences, Beijing, China

E-mail:wylloong@163.com,zy1503245@buaa.edu.cn,yuxuemin@iie.ac.cn

**Abstract:** Unmanned Surface Vehicle (USV) is a new type of intelligent surface craft, and global path planning is the key technology of USV research, which can reflect the intelligent level of USV. In order to solve the problem of global path planning of USV, this paper proposes an improved A* algorithm for sailing cost optimization based on electronic chart. This paper uses the S-57 electronic chart to realize the establishment of the octree grid environment model, and proposes an improved A* algorithm based on sailing safety weight, pilot quantity and path curve smoothing to ensure the safety of the route, reduce the planning time, and improve path smoothness. The simulation results show that the environmental model construction method and the improved A* algorithm can generate safe and reasonable global path.

## 1 Introduction

Unmanned Surface Vehicle (USV) is a new type of intelligent surface craft that is used in military operations, maritime surveillance cruise, and marine environmental monitoring applications, paper [1] summarizes the history, current situation and development trend of USV. Based on the current situation of domestic and foreign research, combined with "US Department of Defense Unmanned Systems Integrated Roadmap "[2] published by the US military and "The Navy Unmanned Surface Vehicle Master Plan "[3] published by the US Navy, the key technologies involved in the research of USV mainly include automatic route generation and path planning technology, autonomous decision-making and collision avoidance technology, water surface object detection and target automatic identification technology and communication technology.

Automatic route generation and path planning technology mainly studies USV global and local path planning, the method of path planning is mainly grid method, artificial potential field method, genetic algorithm method and ant colony algorithm method. Montes modeled Optimum Track Ship Routing (OTSR) for U.S. Navy using a network graph of the Western Pacific Ocean, a binary heap version of Dijkstra's algorithm determines the optimum route [4]. Zhuang presented a search of shortcut Dijkstra algorithm based on electronic chart, which can generate safety and reasonable routes [5].

Electronic chart is a digital map of geographical information and maritime information drawn to suit the needs of navigation. The electronic chart in the USV application is in the exploratory stage at present, mainly used in the path planning and collision avoidance.

This paper argues that autonomous global path planning based on the electronic chart, mainly refers to obtaining necessary information such as sea area geographic information and obstacle information from electronic chart documents, rendering the chart information into an environment model that the USV path planning system recognized, and then proposes an algorithm to implement the global path planning of USV automatically.

## 2  Establishment and Representation of Environmental Model

Radar, camera and other sensor cannot provide the global marine environmental information while USV sails in a wide range of sea, so S-57 electronic chart that can provide detailed and accurate global marine environmental information has become global path planning of USV necessary input, which can guarantee navigation safety.

### 2.1  Electronic chart data extraction

Electronic chart is mainly composed of marine area elements, which can be expressed in detail as submarine terrain, navigation obstacle, navigation sign, port facility and other elements [6]. The S-57 electronic chart standard package format is ISO / IEC 8211 international standard, so this paper uses ISO8211 open source library to resolve all the electronic chart information according to the package structure of S-57 electronic chart [7]. The S-57 electronic chart package format is shown in figure 1 [6].

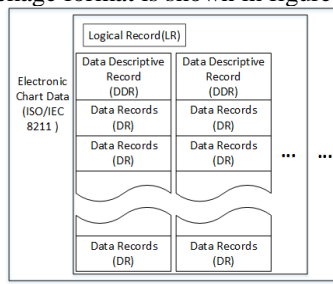

**Fig.1** Package format of S-57 electronic chart

## 2.2 Establishment of environmental model

The establishment of environmental model of USV can be simplified as: USV moves in a limited and arbitrary convex area OS of the sea level, and a limited number of static obstacles $O_i (i=1,2,\cdots,n)$ in OS, obstacle shape and distribution are uncertain. The rasterization of the environmental model is done by filling the OS as a rectangle, and treating the padded area as an obstacle area, and the OS is divided into several meshes of equal size by grid method. This paper judges whether there is an obstacle, such as land and islands in the grid according to information extracted in turn, so as to establish navigable areas and non-navigable areas in units of grids.

The original environmental model and the navigable environmental model are shown in Fig 2 (a) (b), where the shaded portion indicates that it is not navigable.

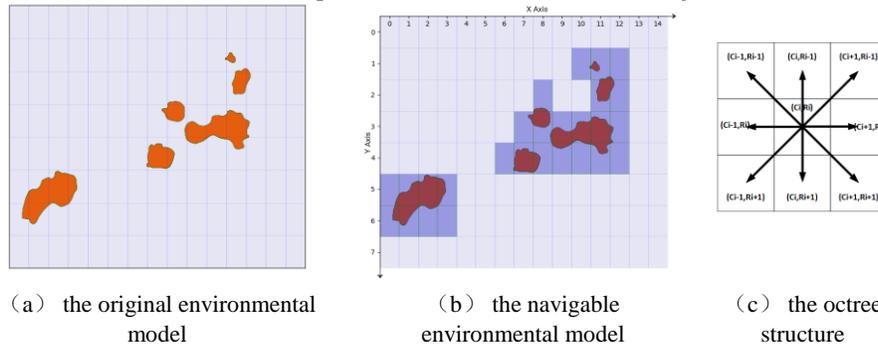

（a）the original environmental model　　（b）the navigable environmental model　　（c）the octree structure

**Fig.2** Display of the environmental models at different stages

In this paper, the octree structure is used as the extension strategy of route search, so there are eight neighboring nodes around the node $i$ as shown in figure 2 (c). USV may deviate from the planned path in the course of avoiding obstacles and temporary mission, and diagonal routes are possible to near non-navigable areas in the octree structure, so this paper considers navigable area near non-navigable area has a potential risk. For the safety of USV, this paper sets sailing safety weight to guide the path planning not to enter potential risk areas. The number of non-navigable grids $n$ in the adjacent eight grids will affect the sailing safety weight of grid ($C_i$, $R_i$) as $w(C_i, R_i) = 1 + \frac{1}{2} * 2^{(n-1)}$.

## 3 The Description and Realization of the Improved A* Algorithm

### 3.1 Realization of the improved A* search algorithm

In the path planning algorithm, A* algorithm is a classical heuristic search algorithm. The basic idea is to use the preset cost function to calculate the value of each adjacent child node that the current node may reach, and select the minimum cost node to join the search space and expand, and so on until the target point is reached [8].

Under the security constraints, the goal of USV path planning optimization is to minimize sailing cost of the planned route. The sailing cost $D(p_i, p_{(i+1)})$ is the straight-line distance between two points multiplied by sailing safety weight, which is given below.

$$D(p_i, p_{(i+1)}) = \sqrt{(x_{pi} - x_{p(i+1)})^2 + (y_{pi} - y_{p(i+1)})^2} * w(C_{(i+1)}, R_{(i+1)}) \tag{1}$$

There is more than one path of the minimum sailing cost sometimes, and they are all explored although only one is needed, and sometimes the planed route automatically is not the reasonable result because of deviation from the straight-line $L_{sg}$ between the starting and target point. Therefore, this paper introduces a pilot quantity P to guide the planed path close to the straight-line $L_{sg}$. Set the starting point as (S.x, S.y), the target point as (G.x, G.y), the current node $i$ as (C.x, C.y), we can get angle $\theta_i$ between the starting-target vector and the i-target vector by calculating vector cross product as:

$$\theta_i = \arcsin\left(\frac{\sqrt{(C.x - G.x)*(S.y - G.y) - (S.x - G.x)*(C.y - G.y)}}{\sqrt{(C.x - G.x)^2 + (C.y - G.y)^2} * \sqrt{(S.x - G.x)^2 + (S.y - G.y)^2}}\right) \tag{2}$$

the pilot quantity $p_{(i)}$ of node $i$ can be expressed as

$$p_{(i)} = 3/(4 - \sin(\theta_i)) \tag{3}$$

In this paper, based on the A* algorithm, the cost function of node $i$ is defined as

$$f_{(i)} = g_{(i)} + h_{(i)} \tag{4}$$

Where $g_{(i)}$ is the sailing cost function from the starting point to the node $i$, $h_{(i)}$ is the heuristic function of current node $i$ to the target point [8], this paper takes the distance multiplied by the pilot quantity as heuristic function $h_{(i)}$, which can ensure the route is optimal because $h_{(i)}$ is not greater than the minimum sailing cost from $i$ to target node.

$$h_{(i)} = \sqrt{(x_{pi} - x_{pG})^2 + (y_{pi} - y_{pG})^2} * p_{(i)} \leq \sqrt{(x_{pi} - x_{pG})^2 + (y_{pi} - y_{pG})^2} \leq \min\left(\sum_{j=i}^{G} D_{(j, j+1)}\right) \tag{5}$$

### 3.2 Path curve smoothing

In the grid environment, if the nodes obtained by the improved A* algorithm are connected in sequence as the planed path of USV, there are sometimes ladder or jagged lines on the route, and it is easy to know that the planned path is not the desired path, so this paper proposes a path curve smoothing method to remove the redundant nodes.

The method of path smoothing is to traverse all the nodes in the planed path, when there is no obstacle on the connection between the node $i$ and $i+2$, then remove redundant node $i+1$. Continue above steps until the connection lines between $i$ and $j$ through the obstacles, then take out 3 consecutive nodes as $P_{j-1}$, $P_j$ and $P_{j+1}$ after node $i$, continue these steps until you have traversed all the route nodes in the path [9].

## 4 Simulation Results

In order to illustrate the effectiveness of environment modeling and the improved A* algorithm, the algorithms mentioned above are simulated. Electronic chart analysis base on C++ language, environment modeling and path planning simulates with python.

For a sea area in the South China Sea (regional range of 18.10° N ~ 18.40° N, 109.35° E ~ 109.85 ° E), given the simulation results of path planning shown in figure 3. Figure (a) is the path planning result of the improved A* algorithm. Figure (b) is the path planning result of A* algorithm. Figure (c) is the path planning result of Dijkstra algorithm, where the solid line is the final planned route, the dotted line is the route without path smoothing.

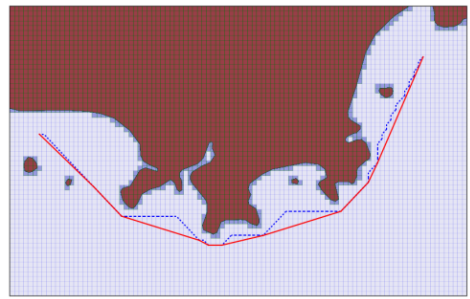

（a）the path planning results of the improved A* algorithm

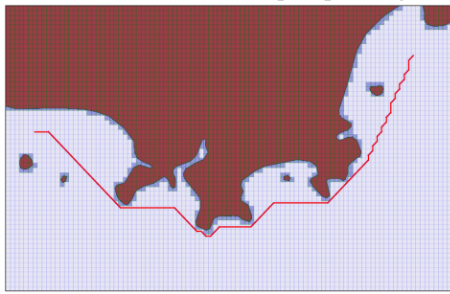 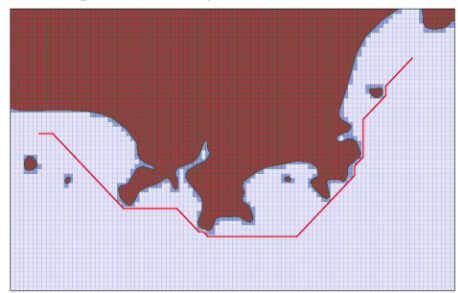

（b）the results of A* algorithm　　　　　（c）the results of Dijkstra algorithm
**Fig.3** Simulation results of several path planning algorithms

Table 1 lists the simulation results of path planning algorithm mentioned above, where number of potential hazards is the total number of non-navigable areas is close to the final route within a grid range, and number of turns is the times of turn in the final route.

**Table.1** Comparison of simulation results of several path planning algorithms

|  | Dijkstra algorithm | A* algorithm | improved A* algorithm |
|---|---|---|---|
| grid accuracy (degrees) | 0.005*0.005 | 0.005*0.005 | 0.005*0.005 |
| sailing distance（Nautical mile） | 35.48 | 35.52 | 34.25 |
| number of route nodes | 97 | 97 | 11 |
| number of nodes traversed | 3186 | 1806 | 2127 |
| number of potential hazards | 32 | 26 | 0 |
| route turn times | 13 | 31 | 6 |

In the simulation results above, we can get from figure (b) (c) that the number of nodes A* algorithm traversed is less than Dijkstra algorithm because Dijkstra algorithm directly searches the global space without considering the target information [10], so that the route planning efficiency of A* algorithm is much higher than Dijkstra algorithm. It can be seen from figure (a) (b) that the improved A* algorithm can improve the safety of route, reduce redundant nodes, improve path smoothness and shorten sailing distance than A* algorithm without improved, and the navigation node layout is more reasonable.

## 5  Conclusion

Considering at the characteristics of USV, this paper proposes an improved A* algorithm for global path planning of USV based on S-57 electronic chart, sailing safety weight, pilot quantity function and path curve smoothing method. This algorithm can quickly and accurately generate an optimal collision-free route between any two possible destinations in a given sea area, which can ensure the safety of the planned route.